\pdfoutput=1
\documentclass[11pt]{article}

\usepackage{acl}

\usepackage{times}
\usepackage{latexsym}
\usepackage{graphicx}

\usepackage[T1]{fontenc}
\usepackage[utf8]{inputenc}

\usepackage{microtype}

\usepackage{inconsolata}

\title{Self-Consistent Decoding for More Factual Open Responses}

\author{Christopher Malon \\
  NEC Laboratories America \\
  \texttt{malon@nec-labs.com} \\
  \And
  Xiaodan Zhu \\
  Queen's University \\
  \texttt{xiaodan.zhu@queensu.ca} \\
}

\begin{document}
\maketitle
\begin{abstract}
Self-consistency has emerged as a powerful method for improving the
accuracy of short answers generated by large language models.
As previously defined, it only concerns the accuracy of a final answer parsed from
generated text.  In this work, we extend the idea to open response generation,
by integrating voting into the decoding method.  Each output sentence is
selected from among multiple samples, conditioning on the previous selections,
based on a simple token overlap score.  We compare this ``Sample \& Select''
method to greedy decoding, beam search, nucleus sampling, and the recently
introduced hallucination avoiding decoders of DoLa, P-CRR, and S-CRR.
We show that Sample \& Select improves factuality by a 30\% relative
margin against these decoders in NLI-based evaluation
on the subsets of CNN/DM and XSum used in the FRANK benchmark,
while maintaining comparable ROUGE-1 F1 scores against reference summaries.
We collect human verifications of the generated summaries,
confirming the factual superiority of our method.
\end{abstract}

\section{Introduction}

A large language model (LLM) generates output text by predicting
probabilities for each token, conditioned on a prompt and
previous tokens in the output, and using a decoding strategy such as
greedy decoding, beam search, or nucleus sampling to select tokens at each
position and assemble an output string.
Like any abstractive generative model,
LLM's have a risk of outputting information that is false or unsupported
with respect to a context or closed book knowledge, commonly called
{\em hallucination} \citep{kryscinski-etal-2020-evaluating, fabbri-etal-2021-summeval, manakul-etal-2023-selfcheckgpt}.

Recent work has observed that if multiple responses are sampled from
a large language model, they are likely to hallucinate different
values for each detail
\citep{manakul-etal-2023-selfcheckgpt, selfconsistency}.
On the other hand, details that multiple sampled responses share in common
are likely to be the truth.

In cases where a short answer
can be easily parsed from a generated response, {\em self consistency}
\citep{selfconsistency}
uses the idea of voting (using exact matches of the parsed short answer)
to select a better answer more reliably than
using an individual sample.  However, this procedure does not apply to tasks
where the full response is needed and not just an extracted short answer.

Our main contribution is a novel decoding mechanism which can be applied to any
LLM task, which samples multiple text outputs from the LLM and selects the
most consistent choice for each sentence, reflecting the commonality
in the responses.  We compare our decoding technique to several baselines
in automatic factuality metrics, applied to Llama 2 (chat) and Mistral.
We also contribute human evaluations, including a coarse classification
of errors.
In the accompanying code, we provide an implementation of our method
and the baselines, along with the human evaluation
dataset.\footnote{{\tt https://github.com/cdmalon/selfconsistent}}

\begin{figure*}[htb]
\begin{center}
\includegraphics[height=3in,width=6in]{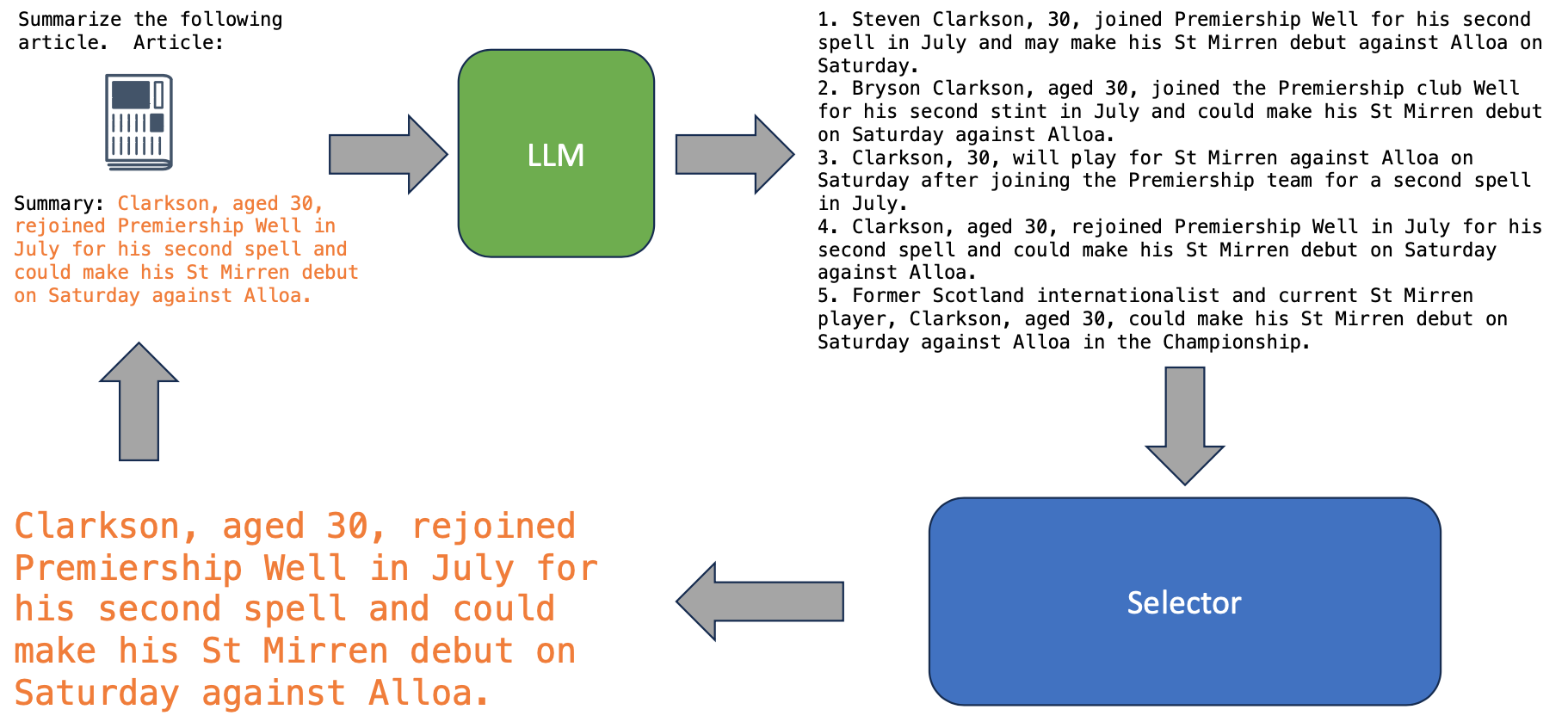}
\caption{Sampling and selecting sentences iteratively.}
\label{fig:iteration}
\end{center}
\end{figure*}

\section{Related work}

Recent research has begun to introduce decoding strategies to avoid
hallucinations.
Faithfulness-Aware Decoding and Context-Aware Decoding assume that
generated text should be supported by a provided
context \citep{wan-etal-2023-faithfulness, shi2023trusting}.
Decoding by Contrasting Layers (DoLa) and Certainty-Based Response Ranking (CRR)
apply more generally to the case where responses are not to be limited to
information in the context \citep{dola, pcrr}.
We are interested in this setting and consider these techniques
as baselines.

SelfCheckGPT \citep{manakul-etal-2023-selfcheckgpt} introduces a technique for scoring
the hallucination level of an open-ended generation using multiple sampled
responses.  Although it does not provide a decoding technique for making
a more factual generation, we instantiate our decoding technique
with a score they introduce as a further baseline, and show that our score
performs better.

\section{Method: Sample and Select}

Given a large language model (LLM) and prompt text, generation of output text
proceeds as follows.  Using nucleus sampling \citep{nucleus}
with $p = .9$, $n = 5$
sampled outputs are decoded from the LLM until the end of
one sentence in each.

We count the number of other sampled sentences from the LLM that matched
each token of each sampled output, and score each sentence by the
average count per token.  Formally, if sampled
sentences $s_1, \ldots, s_n$ consist of words $w_i^j$, $j = 1 \ldots m_i$,
the score of sample $i$ is
$$\frac{1}{m_i} \sum_{j=1}^{m_i} \sum_{k=1}^n 1_{w_i^j \in s_k}$$
where $1_{w_i^j \in s_k}$ is 1 if token $w_i^j$ is in sentence $s_k$ and
0 if not.
This score embodies the intuition that details are represented by tokens,
and that each trustworthy detail should occur in a large number of other
samples.
In the case where no token is repeated in a sample, this score is proportional
to a sum
of ROUGE-1 precisions with the other samples, but unlike ROUGE-1, our score
is not concerned with comparing numbers of repetitions.
\citet{manakul-etal-2023-selfcheckgpt} found that another kind of unigram score
(which we compare as a baseline) outperformed scores using higher $n$-grams
or BERTscore \citep{bertscore} in detecting hallucinations, supporting
our focus on token-level details.

We check each sample for grammaticality by the existence of both
a subject and a verb, by computing parts of speech and dependency parsing,
comparing to a set of designated tags listed in Appendix~\ref{sec:gram}.
Scores of samples that fail the check are set to zero.
Generation may be aborted if no sampled generalization outputs survive.
The sampled output with the highest score
is added to the final output and the prompt.

Additional sentences are added to the final output iteratively
(Figure~\ref{fig:iteration}),
collecting new samples conditioning on the prompt including
previously chosen sentences.  If any sample ends after the current sentence,
generation is stopped after selection and the final output is returned.

The time to compute token overlap scores is negligible, so the method
could run as efficiently as collecting multiple samples from nucleus sampling,
if the implementation would stop the generation as soon as the end of a
sentence is reached.

\section{Experiment Setup}

\begin{table*}[htb]
\footnotesize \resizebox{1\linewidth}{!}{ \parbox{1\linewidth}{
\begin{center}
\begin{tabular}{lcccccccc}
\hline
& \multicolumn{4}{c}{Llama 2} & \multicolumn{4}{c}{Mistral} \\
Method & \multicolumn{2}{c}{SummaC} & ROUGE-1 & QAFE &
\multicolumn{2}{c}{SummaC} & ROUGE-1 & QAFE \\
& ZS & Conv & F1 & F1 & ZS & Conv & F1 & F1 \\
\hline
Greedy & .447 & .409 & .338 & .595 
& .423 & .401 & .311 & .552 \\
Nucleus & .448 & .420 & .341 & .599
& .412 & .378 & .313 & .531 \\
Beam & .474 & .444 & {\bf .345} & .637 
& .447 & .430 & .310 & {\bf .608} \\
\hline
P-CRR & .464 & .466 & .338 & .628 
& .438 & .409 & .313 & .563 \\
S-CRR & .453 & .418 & .334 & .590 
& .441 & .395 & .308 & .527 \\
DoLa & .436 & .407 & .336 & .586 
& --- & --- & --- & --- \\
SelfCheckGPT & .550 & .521 & .334 & .640 
& .417 & .396 & .313 & .534 \\
\hline
Independent & .510 & .457 & .331 & .596 
& .476 & .431 & .316 & .537 \\
Sample \& Select & {\bf .619} & {\bf .575} & .337 & {\bf .642}
& {\bf .505} & {\bf .452} & {\bf .327} & .557 \\
\hline
\end{tabular}
\caption{Automatic evaluations on CNN/Daily Mail.}
\label{tbl:cnnauto}
\end{center}
}}
\end{table*}

\begin{table*}[htb]
\footnotesize \resizebox{1\linewidth}{!}{ \parbox{1\linewidth}{
\begin{center}
\begin{tabular}{lcccccccc}
\hline
& \multicolumn{4}{c}{Llama 2} & \multicolumn{4}{c}{Mistral} \\
Method & \multicolumn{2}{c}{SummaC} & ROUGE-1 & QAFE &
\multicolumn{2}{c}{SummaC} & ROUGE-1 & QAFE \\
& ZS & Conv & F1 & F1 & ZS & Conv & F1 & F1 \\
\hline
Greedy & .356 & .364 & .089 & .580
& .413 & .402 & .075 & .575 \\
Nucleus & .378 & .390 & .089 & .572
& .387 & .376 & .075 & .538 \\
Beam & .386 & .395 & .089 & {\bf .616}
& .420 & .414 & .072 & {\bf .621} \\
\hline
P-CRR & .380 & .418 & .086 & .607
& .409 & .386 & .074 & .576 \\
S-CRR & .358 & .368 & .084 & .576
& .394 & .377 & .073 & .536 \\
DoLa & .339 & .360 & .088 & .560
& --- & --- & --- & --- \\
SelfCheckGPT & .425 & .428 & .085 & .585
& .382 & .379 & .075 & .543 \\
\hline
Independent & .394 & .390 & .085 & .568
& .451 & .414 & .081 & .557 \\
Sample \& Select & {\bf .522} & {\bf .488} & {\bf .090} & .600
& {\bf .474} & {\bf .435} & {\bf .080} & .559 \\
\hline
\end{tabular}
\caption{Automatic evaluations on XSum.}
\label{tbl:xsumauto}
\end{center}
}}
\end{table*}

\begin{table*}[htb]
\footnotesize \resizebox{1\linewidth}{!}{ \parbox{1\linewidth}{
\begin{center}
\begin{tabular}{llcccccc}
\hline
Data & Method & No Error & Contradiction & Meaningless & Other & Pronoun & Unsupported \\
\hline
CNN/DM & Beam & .690 & .041 & .174 & {\bf .006} & .003 & .085 \\
& SelfCheckGPT & .713 & {\bf .028} & .196 & .016 & {\bf .000} & .047 \\
& Independent & .686 & .040 & {\bf .140} & .012 & {\bf .000} & .121 \\
& Sample \& Select & {\bf .747} & .035 & .147 & .019 & .005 & {\bf .046} \\
\hline
XSum & Beam & .739 & {\bf .008} & .171 & {\bf .006} & {\bf .000} & .077 \\
& SelfCheckGPT & .638 & .033 & .202 & .036 & .008 & .083 \\
& Independent & .732 & .021 & {\bf .130} & .021 & .006 & .091 \\
& Sample \& Select & {\bf .741} & .025 & .145 & .027 & .002 & {\bf .061} \\
\hline
\end{tabular}
\caption{Human evaluations on Llama 2 summaries.}
\label{tbl:human}
\end{center}
}}
\end{table*}

We evaluate Llama 2 chat (13 billion)
\citep{llama} and Mistral-7B-Instruct-v0.2 \citep{mistral}
 at zero-shot summarization
on English news articles
from the CNN/DM \citep{cnn} and XSum \citep{narayan-etal-2018-dont} datasets,
comparing the results obtained based on different decoding methods.
From each dataset, we use the subset of 175 documents appearing
in the test subset of FRANK \citep{pagnoni-etal-2021-understanding}, a benchmark in which
human and automatic verifications of previous systems have been collected.
To fix sentence segmentation issues, the articles are cleaned with
regular expressions before summarization
(see Appendix~\ref{sec:cleanup}), but not when computing the metrics.

For automatic factuality evaluation, we use
SummaC \citep{laban-etal-2022-summac} as a
natural language inference based measure, and QAFactEval (QAFE) 
\citep{fabbri-etal-2022-qafacteval} as a question answering based measure.
% QAFE performs better than QuestEval on FactCC, per 2305.14069
We also report ROUGE-1 F1
scores\footnote{Huggingface Datasets 2.3.2}
\citep{lin-2004-rouge}.
SummaC is run in the zero-shot (ZS) and convolutional (Conv) variants
at sentence granularity, using the ``vitc'' model.
For QAFactEval, we report the average F1 score.

For human factuality evaluation, we randomly choose 50 articles
from the FRANK test set and evaluate the four systems with the
highest SummaCZS scores.
We solicit three human evaluations
of each sentence of each system generated summary from a trusted pool
of Amazon Mechanical Turk crowdworkers (instructions in
Appendix~\ref{sec:label} and recruitment protocol in Appendix~\ref{sec:recruiting}).  Each system's score is the average over
all worker ratings of its generated sentences.
Over all worker ratings, there is 94.3\% agreement with
majority labels (6.8\% false positive, 5.3\% false negative).

As baselines, we consider {\bf greedy} decoding,
{\bf beam} search with five beams,
and {\bf nucleus} sampling ($p = .9$) \citep{nucleus}.
Furthermore, we test four recent decoding techniques designed to reduce
hallucinations:

\begin{enumerate}
\item {\bf DoLa} \citep{dola} modifies the next token distribution by
contrasting the difference in logits obtained from a projections of late
and early layers of the LLM.  Following the paper, the
even numbered layers in the first half of the layers of the transformer
are used as candidate premature layers for dynamic premature layer
selection, contrasted with layer 40.  The official implementation works
with Llama but not Mistral.

\item Probabilistic certainty-based response ranking ({\bf P-CRR}) \citep{pcrr}
ranks responses using the mean log probability of the entire sequence.
We implement P-CRR ourselves.

\item Semantic certainty-response ranking ({\bf S-CRR}) \citep{pcrr} ranks responses using an entailment-based agreement score.  We implement S-CRR,
following the paper to judge entailment using a Roberta-Large model trained on
Adversarial NLI (ANLI) \citep{nie-etal-2020-adversarial}.

\item {\bf SelfCheckGPT} \citep{manakul-etal-2023-selfcheckgpt}
uses sampling to detect hallucination, similar to
our approach.  The paper does not propose a decoding approach
but only scores sentences according to likelihood of hallucination.
We extend this to a decoding method by choosing, for each sentence,
the sample with the lowest hallucination score, re-conditioning on the
chosen sentences.  This extension parallels our method.  The paper proposes
several possible scores, some of which are too slow for practical use
(their QA-based score took 4.5 minutes for one example) or effectively
call the factuality evaluator to choose the output.
We use the efficient unigram score,
based on negative log likelihood of tokens based on their occurence
in other responses, which performs well in their evaluation.
\end{enumerate}

Our method is referred to as {\bf sample \& select} in the tables.
As an ablation, we consider {\bf independent}, which decodes entire responses
and selects each sentence in the same way, without reconditioning later
sentences on the finally chosen sentences.

\section{Results and Discussion}

Tables~\ref{tbl:cnnauto} and~\ref{tbl:xsumauto} show automatic evaluations
of all systems.  Our system leads both SummaC metrics by large margins
on all models.
The comparison with {\em independent} shows that conditioning later
sentences on finally chosen samples is critical.  This advantage is
included in our design of the SelfCheckGPT baseline, which performs
next best on Llama 2 and whose gap reflects our improvement in the scoring
function.
We gain at least 30\% (relative) in SummaCZS with Llama 2
over all baselines behind SelfCheckGPT.

The strong performance of beam search among the standard decoding
techniques was already noted by \citet{wan-etal-2023-faithfulness}, and
beam search occasionally beats Sample \& Select in the QAFactEval metric.
We further explore this phenomenon in Appendix~\ref{sec:metrics}, where we show
the errors that are corrected might not always concern factoids for which
questions are generated.
Although ROUGE requires the original reference summaries and is not strongly factuality oriented, we include it to show that our
technique remains comparable to scores of other techniques, while gaining
factuality.

Humans judge Sample \& Select to have the highest factual accuracy and
lowest number of unsupported sentences, as shown in Table~\ref{tbl:human}.
However, our technique seems not to avert the
generation of meaningless sentences or pronoun errors.  We instructed the
graders to label ``Sure!'' at the head of a summary as a meaningless
sentence, and this appears in the majority of Llama 2 13B summaries with
any decoding method.

\section{Conclusion}

Sampling multiple responses one sentence at a time, and choosing one with
a simple, efficient token overlap consistency score, can significantly
increase the factuality of a generated response.
Other hallucination estimates may be effectively used in place of
token overlap, but the unigram score described in SelfCheckGPT was at least
10\% worse in NLI-based evaluation.   Conditioning each sentence on
the previously chosen sentence is essential; if further sentences are generated
conditioned on an unchosen sample, factuality improves by less.

Although it is beyond the scope of
the current paper, Sample \& Select can apply beyond summarization to other
generation tasks.  We hope future researchers will find this technique
helpful and easy to drop into their generation pipeline.

\section*{Limitations}

The proposed decoding method is sensitive to sentence splitting
and tokenization.  The scoring method involves a heuristic to choose
more reliable samples, and further evaluation is needed to confirm that
output chosen in this way enhances factuality in more settings, particularly
on tasks beyond summarization.  The rules of the grammaticality filter were
designed specifically for English language text.

\bibliography{anthology,select4}

\begin{thebibliography}{21}
\expandafter\ifx\csname natexlab\endcsname\relax\def\natexlab#1{#1}\fi

\bibitem[{Bird and Loper(2004)}]{bird-loper-2004-nltk}
Steven Bird and Edward Loper. 2004.
\newblock \href {https://aclanthology.org/P04-3031} {{NLTK}: The natural
  language toolkit}.
\newblock In \emph{Proceedings of the {ACL} Interactive Poster and
  Demonstration Sessions}, pages 214--217, Barcelona, Spain. Association for
  Computational Linguistics.

\bibitem[{Chuang et~al.(2023)Chuang, Xie, Luo, Kim, Glass, and He}]{dola}
Yung-Sung Chuang, Yujia Xie, Hongyin Luo, Yoon Kim, James Glass, and Pengcheng
  He. 2023.
\newblock \href {https://arxiv.org/abs/2309.03883} {{D}o{L}a: Decoding by
  contrasting layers improves factuality in large language models}.
\newblock In \emph{arXiv}, 2309.03883.

\bibitem[{Fabbri et~al.(2022)Fabbri, Wu, Liu, and
  Xiong}]{fabbri-etal-2022-qafacteval}
Alexander Fabbri, Chien-Sheng Wu, Wenhao Liu, and Caiming Xiong. 2022.
\newblock \href {https://doi.org/10.18653/v1/2022.naacl-main.187}
  {{QAF}act{E}val: Improved {QA}-based factual consistency evaluation for
  summarization}.
\newblock In \emph{Proceedings of the 2022 Conference of the North American
  Chapter of the Association for Computational Linguistics: Human Language
  Technologies}, pages 2587--2601, Seattle, United States. Association for
  Computational Linguistics.

\bibitem[{Fabbri et~al.(2021)Fabbri, Kry{\'s}ci{\'n}ski, McCann, Xiong, Socher,
  and Radev}]{fabbri-etal-2021-summeval}
Alexander~R. Fabbri, Wojciech Kry{\'s}ci{\'n}ski, Bryan McCann, Caiming Xiong,
  Richard Socher, and Dragomir Radev. 2021.
\newblock \href {https://doi.org/10.1162/tacl_a_00373} {{S}umm{E}val:
  Re-evaluating summarization evaluation}.
\newblock \emph{Transactions of the Association for Computational Linguistics},
  9:391--409.

\bibitem[{Hermann et~al.(2015)Hermann, Kocisky, Grefenstette, Espeholt, Kay,
  Suleyman, and Blunsom}]{cnn}
Karl~Moritz Hermann, Tomas Kocisky, Edward Grefenstette, Lasse Espeholt, Will
  Kay, Mustafa Suleyman, and Phil Blunsom. 2015.
\newblock \href
  {https://proceedings.neurips.cc/paper_files/paper/2015/file/afdec7005cc9f14302cd0474fd0f3c96-Paper.pdf}
  {Teaching machines to read and comprehend}.
\newblock In \emph{Advances in Neural Information Processing Systems},
  volume~28. Curran Associates, Inc.

\bibitem[{Holtzman et~al.(2020)Holtzman, Buys, Du, Forbes, and Choi}]{nucleus}
Ari Holtzman, Jan Buys, Li~Du, Maxwell Forbes, and Yejin Choi. 2020.
\newblock \href {https://openreview.net/forum?id=rygGQyrFvH} {The curious case
  of neural text degeneration}.
\newblock In \emph{International Conference on Learning Representations}.

\bibitem[{Honnibal et~al.(2020)Honnibal, Montani, Van~Landeghem, and
  Boyd}]{spacy}
Matthew Honnibal, Ines Montani, Sofie Van~Landeghem, and Adriane Boyd. 2020.
\newblock \href {https://doi.org/10.5281/zenodo.1212303} {{spaCy:
  Industrial-strength Natural Language Processing in Python}}.

\bibitem[{Jiang et~al.(2023)Jiang, Sablayrolles, Mensch, Bamford, Chaplot,
  de~las Casas, Bressand, Lengyel, Lample, Saulnier, Lavaud, Lachaux, Stock,
  Scao, Lavril, Wang, Lacroix, and Sayed}]{mistral}
Albert~Q. Jiang, Alexandre Sablayrolles, Arthur Mensch, Chris Bamford,
  Devendra~Singh Chaplot, Diego de~las Casas, Florian Bressand, Gianna Lengyel,
  Guillaume Lample, Lucile Saulnier, Lélio~Renard Lavaud, Marie-Anne Lachaux,
  Pierre Stock, Teven~Le Scao, Thibaut Lavril, Thomas Wang, Timothée Lacroix,
  and William~El Sayed. 2023.
\newblock \href {https://arxiv.org/abs/2310.06825} {Mistral 7b}.
\newblock In \emph{arXiv}, 2310.06825.

\bibitem[{Kryscinski et~al.(2020)Kryscinski, McCann, Xiong, and
  Socher}]{kryscinski-etal-2020-evaluating}
Wojciech Kryscinski, Bryan McCann, Caiming Xiong, and Richard Socher. 2020.
\newblock \href {https://doi.org/10.18653/v1/2020.emnlp-main.750} {Evaluating
  the factual consistency of abstractive text summarization}.
\newblock In \emph{Proceedings of the 2020 Conference on Empirical Methods in
  Natural Language Processing (EMNLP)}, pages 9332--9346, Online. Association
  for Computational Linguistics.

\bibitem[{Laban et~al.(2022)Laban, Schnabel, Bennett, and
  Hearst}]{laban-etal-2022-summac}
Philippe Laban, Tobias Schnabel, Paul~N. Bennett, and Marti~A. Hearst. 2022.
\newblock \href {https://doi.org/10.1162/tacl_a_00453} {{S}umma{C}: Re-visiting
  {NLI}-based models for inconsistency detection in summarization}.
\newblock \emph{Transactions of the Association for Computational Linguistics},
  10:163--177.

\bibitem[{Lin(2004)}]{lin-2004-rouge}
Chin-Yew Lin. 2004.
\newblock \href {https://aclanthology.org/W04-1013} {{ROUGE}: A package for
  automatic evaluation of summaries}.
\newblock In \emph{Text Summarization Branches Out}, pages 74--81, Barcelona,
  Spain. Association for Computational Linguistics.

\bibitem[{Manakul et~al.(2023)Manakul, Liusie, and
  Gales}]{manakul-etal-2023-selfcheckgpt}
Potsawee Manakul, Adian Liusie, and Mark Gales. 2023.
\newblock \href {https://aclanthology.org/2023.emnlp-main.557}
  {{S}elf{C}heck{GPT}: Zero-resource black-box hallucination detection for
  generative large language models}.
\newblock In \emph{Proceedings of the 2023 Conference on Empirical Methods in
  Natural Language Processing}, pages 9004--9017, Singapore. Association for
  Computational Linguistics.

\bibitem[{Narayan et~al.(2018)Narayan, Cohen, and
  Lapata}]{narayan-etal-2018-dont}
Shashi Narayan, Shay~B. Cohen, and Mirella Lapata. 2018.
\newblock \href {https://doi.org/10.18653/v1/D18-1206} {Don{'}t give me the
  details, just the summary! topic-aware convolutional neural networks for
  extreme summarization}.
\newblock In \emph{Proceedings of the 2018 Conference on Empirical Methods in
  Natural Language Processing}, pages 1797--1807, Brussels, Belgium.
  Association for Computational Linguistics.

\bibitem[{Nie et~al.(2020)Nie, Williams, Dinan, Bansal, Weston, and
  Kiela}]{nie-etal-2020-adversarial}
Yixin Nie, Adina Williams, Emily Dinan, Mohit Bansal, Jason Weston, and Douwe
  Kiela. 2020.
\newblock \href {https://doi.org/10.18653/v1/2020.acl-main.441} {Adversarial
  {NLI}: A new benchmark for natural language understanding}.
\newblock In \emph{Proceedings of the 58th Annual Meeting of the Association
  for Computational Linguistics}, pages 4885--4901, Online. Association for
  Computational Linguistics.

\bibitem[{Pagnoni et~al.(2021)Pagnoni, Balachandran, and
  Tsvetkov}]{pagnoni-etal-2021-understanding}
Artidoro Pagnoni, Vidhisha Balachandran, and Yulia Tsvetkov. 2021.
\newblock \href {https://doi.org/10.18653/v1/2021.naacl-main.383}
  {Understanding factuality in abstractive summarization with {FRANK}: A
  benchmark for factuality metrics}.
\newblock In \emph{Proceedings of the 2021 Conference of the North American
  Chapter of the Association for Computational Linguistics: Human Language
  Technologies}, pages 4812--4829, Online. Association for Computational
  Linguistics.

\bibitem[{Shi et~al.(2023)Shi, Han, Lewis, Tsvetkov, Zettlemoyer, and tau
  Yih}]{shi2023trusting}
Weijia Shi, Xiaochuang Han, Mike Lewis, Yulia Tsvetkov, Luke Zettlemoyer, and
  Scott~Wen tau Yih. 2023.
\newblock \href {https://arxiv.org/abs/2305.14739} {Trusting your evidence:
  Hallucinate less with context-aware decoding}.
\newblock In \emph{arXiv}, 2305.14739.

\bibitem[{Touvron et~al.(2023)Touvron, Martin, Stone, Albert, Almahairi,
  Babaei, Bashlykov, Batra, Bhargava, Bhosale, Bikel, Blecher, Ferrer, Chen,
  Cucurull, Esiobu, Fernandes, Fu, Fu, Fuller, Gao, Goswami, Goyal, Hartshorn,
  Hosseini, Hou, Inan, Kardas, Kerkez, Khabsa, Kloumann, Korenev, Koura,
  Lachaux, Lavril, Lee, Liskovich, Lu, Mao, Martinet, Mihaylov, Mishra,
  Molybog, Nie, Poulton, Reizenstein, Rungta, Saladi, Schelten, Silva, Smith,
  Subramanian, Tan, Tang, Taylor, Williams, Kuan, Xu, Yan, Zarov, Zhang, Fan,
  Kambadur, Narang, Rodriguez, Stojnic, Edunov, and Scialom}]{llama}
Hugo Touvron, Louis Martin, Kevin Stone, Peter Albert, Amjad Almahairi, Yasmine
  Babaei, Nikolay Bashlykov, Soumya Batra, Prajjwal Bhargava, Shruti Bhosale,
  Dan Bikel, Lukas Blecher, Cristian~Canton Ferrer, Moya Chen, Guillem
  Cucurull, David Esiobu, Jude Fernandes, Jeremy Fu, Wenyin Fu, Brian Fuller,
  Cynthia Gao, Vedanuj Goswami, Naman Goyal, Anthony Hartshorn, Saghar
  Hosseini, Rui Hou, Hakan Inan, Marcin Kardas, Viktor Kerkez, Madian Khabsa,
  Isabel Kloumann, Artem Korenev, Punit~Singh Koura, Marie-Anne Lachaux,
  Thibaut Lavril, Jenya Lee, Diana Liskovich, Yinghai Lu, Yuning Mao, Xavier
  Martinet, Todor Mihaylov, Pushkar Mishra, Igor Molybog, Yixin Nie, Andrew
  Poulton, Jeremy Reizenstein, Rashi Rungta, Kalyan Saladi, Alan Schelten, Ruan
  Silva, Eric~Michael Smith, Ranjan Subramanian, Xiaoqing~Ellen Tan, Binh Tang,
  Ross Taylor, Adina Williams, Jian~Xiang Kuan, Puxin Xu, Zheng Yan, Iliyan
  Zarov, Yuchen Zhang, Angela Fan, Melanie Kambadur, Sharan Narang, Aurelien
  Rodriguez, Robert Stojnic, Sergey Edunov, and Thomas Scialom. 2023.
\newblock \href {http://arxiv.org/abs/2307.09288} {Llama 2: Open foundation and
  fine-tuned chat models}.
\newblock In \emph{arXiv}, 2307.09288.

\bibitem[{Wan et~al.(2023{\natexlab{a}})Wan, Liu, McKeown, Dreyer, and
  Bansal}]{wan-etal-2023-faithfulness}
David Wan, Mengwen Liu, Kathleen McKeown, Markus Dreyer, and Mohit Bansal.
  2023{\natexlab{a}}.
\newblock \href {https://doi.org/10.18653/v1/2023.eacl-main.210}
  {Faithfulness-aware decoding strategies for abstractive summarization}.
\newblock In \emph{Proceedings of the 17th Conference of the European Chapter
  of the Association for Computational Linguistics}, pages 2864--2880,
  Dubrovnik, Croatia. Association for Computational Linguistics.

\bibitem[{Wan et~al.(2023{\natexlab{b}})Wan, Wu, Xu, and Sengamedu}]{pcrr}
Yixin Wan, Fanyou Wu, Weijie Xu, and Srinivasan~H. Sengamedu.
  2023{\natexlab{b}}.
\newblock \href {https://arxiv.org/abs/2310.18794} {Sequence-level certainty
  reduces hallucination in knowledge-grounded dialogue generation}.
\newblock In \emph{arXiv}, 2310.18794.

\bibitem[{Wang et~al.(2023)Wang, Wei, Schuurmans, Le, Chi, Narang, Chowdhery,
  and Zhou}]{selfconsistency}
Xuezhi Wang, Jason Wei, Dale Schuurmans, Quoc~V Le, Ed~H. Chi, Sharan Narang,
  Aakanksha Chowdhery, and Denny Zhou. 2023.
\newblock \href {https://openreview.net/forum?id=1PL1NIMMrw} {Self-consistency
  improves chain of thought reasoning in language models}.
\newblock In \emph{The Eleventh International Conference on Learning
  Representations}.

\bibitem[{Zhang* et~al.(2020)Zhang*, Kishore*, Wu*, Weinberger, and
  Artzi}]{bertscore}
Tianyi Zhang*, Varsha Kishore*, Felix Wu*, Kilian~Q. Weinberger, and Yoav
  Artzi. 2020.
\newblock \href {https://openreview.net/forum?id=SkeHuCVFDr} {Bertscore:
  Evaluating text generation with bert}.
\newblock In \emph{International Conference on Learning Representations}.

\end{thebibliography}

% \pagebreak

\appendix

\section{Grammar Checking}

\label{sec:gram}

The goal of grammar checking is to reject incomplete sentence fragments.
Sentences were tokenized using the \verb+sentence_tokenize+ function
in the \verb+tokenize+ module of NLTK 3.8.1 \citep{bird-loper-2004-nltk}.

The \verb+en_core_web_sm+ models of Spacy 3.7.0 \citep{spacy} were used
to run part of speech tagging and dependency parsing on each sentence.
A sentence was accepted as grammatical if a subject and a verb was found.
We considered that a subject was found if any token in the sentence
was labeled with the ``nsubj,'' ``nsubjpass,'' or ``expl'' dependency
relations.  We considered that a verb was found if any token was tagged
with the part of speech ``VBZ,'' ``VBD,'' or ``VBP,'' 
or the dependency relations ``aux'' or ``auxpass.''

\section{Article Cleanup}

\label{sec:cleanup}

Because missing whitespace, run-together words, and
social media buttons confounded proper sentence segmentation
with NLTK \verb+sentence_tokenize+,
we applied the following Python regular expression substitutions
to the article text.
These rules were inspired by our observations on the validation sets.

\begin{verbatim}
document = re.sub(r"\.([a-zA-Z])",
 ". \\1", document)
document = re.sub(r"([a-z])([A-Z])",
 "\\1 \\2", document)
document = re.sub(r"Share this with
  Email Facebook Messenger Messenger
  Twitter Pinterest Whats App Linked
  In Copy this link", "", document)
\end{verbatim}

\section{Labeling Protocol}

\label{sec:label}

The Turks were given the following instructions to label
each sentence of a system generated summary, given an article:

\begin{itemize}
\item Read the text carefully.
\item You're given a news article.  Don't let it bother you if there is some repetition or sentences without periods in the article text - these were just picture captions or callouts that interrupted the flow of the main text.
\item A machine-generated summary of the article is split into sentences.
\item Decide whether all details of each summary sentence are fully supported
by the article, or not.
\item Note that some words in the passage are replaced with ``UNK'',
a placeholder for unknown words.  These words are not errors.
\item If there are no errors, select ``No Error.''
\item If there is an error, select ``Error.''  Possible reasons for an
error include:
\begin{itemize}
\item {\bf Contradicts article}: Use this for errors that are direct
contradictions.  This could result from the wrong entities, the wrong
relations, or wrongly attached adjectives, adverbs, or other description.
\item {\bf Adds unsupported information}: If not a direct contradiction,
any new information (even if true) that does not appear in the original
article should be considered as this type.  For example, if ``Bob Jones''
is mentioned in the summary but the article only refers to
``Spokesman Jones'' without a first name, this is a new, unsupported
detail.
\item {\bf Meaningless}: Minor grammatical errors may be ignored, but use
this for errors so serious that the meaning of the sentence is
unclear, such as ``The vaccine accepted have already started.''
This also applies to sentences that appear totally out of context,
like ``Sure!''
\item {\bf Pronoun reference}: This refers to cases where a word like
``it'' or ``he'' in the summary refers to the wrong entity (or hasn't been
specified) in the summary, but the statement would be correct if the
pronoun referred to a different entity.
\item {\bf Other}
\end{itemize}
\end{itemize}

Each summary was presented to three crowdworkers for labeling.
Our error types simplify the typology of FRANK because fine-grained agreement
about error types was low on FRANK.

\section{Crowdworker Recruitment}

\label{sec:recruiting}

For every four system summaries in our main labeling task,
we added one alertness test from the original FRANK dataset.
This test was an old system summary
in which one of the sentences had a unanimous error or non-error label.
The purpose of the test was to check whether our crowdworker agreed
with the unanimous label.  The fine-grained type of the error was
not considered.

Our first qualification was a seven-question test, labeling individual
sentences from FRANK on which the original annotators had unanimous
agreement.  Prospective workers had to label ``error'' or ``no error''
correctly on six of the seven problems, but did not have to choose
the error type.  Additionally, crowdworkers had to be in the United States,
Great Britain, or Australia, have a 90\% HIT approval rating, and at
least 1000 approved HITs.

Qualified workers were allowed to participate in an open round of
labeling, where anyone who met the qualifications could participate.
This round used 25 articles different from the ones on which we
reported final test results.  From this pool, we granted an exclusive
qualification to the workers who passed a high fraction of the alertness
tests, did not submit answers too quickly, and had low
false positive and false negative rates with respect to
majority labels (error or non-error, not considering the type).
The exact thresholds are secret, to encourage good
behavior by Turks.  Monitoring behavior on an open round was necessary
even after the qualification test, because the majority of workers
in the open round were careless, aiming to collect rewards as fast as
possible, and did not meet these qualification criteria.

We also granted the exclusive qualification to more workers
whose integrity was demonstrated on previous labeling tests by the authors.
These workers still had to pass the qualification test.

Then we held closed rounds of labeling, only open to those with the exclusive
qualification.
We continually monitored the criteria for the exclusive
qualification on each batch.  One labeler turned rogue and submitted
highly inaccurate (with respect to majority) labels much too quickly,
and his qualification was revoked.  Labels by this worker were excluded
from the final dataset, leaving some summaries with only two
crowdworker annotations.

The workers were paid \$1.20 per system summary, based on an estimate
that a typical summary would require four minutes to grade,
for a targeted compensation of \$18 per hour.  This is above
minimum wage anywhere in the United States as of December 12, 2023.\footnote{{\em Minimum Wage in the United States}, Wikipedia.}
We informed the workers that their annotations were used to make
published papers and high quality datasets.

\section{Metrics and Human Judgments}
\label{sec:metrics}

In the set of summaries judged by our crowdworkers, we computed
QAFE and SummaC scores on a sentence basis, and compared to
crowdworkers' judgments of each sentence as ``NoError'' versus
any type of error (hallucination).  Both SummaCZS and QAFE have high AUC scores
(.82 and .85) in predicting the majority decision.

Focusing on cases where SummaCZS and QAFE disagree, at threshold 0.5,
we find 37 hallucination
sentences with high QAFE and low SummaCZS score,
and only 31 with high SummaCZS and low QAFE.

Qualitatively, we have the following findings.

{\bf QAFE may fail to catch an error when different
phrases of the article are improperly linked together.}
In the first of the 37 examples, ``He missed the Premier League since 2006 and felt emotional when he scored the goal, comparing it to scoring in the World Cup,''
the player actually missed the World Cup in 2006 and missed the Premier League
since some unspecified time.  QAFE's generated questions combine many
details from the generated summary, such as,
``What league did he miss since 2006 and feel emotional when he scored
        the goal, comparing it to scoring in the World Cup?''
and ``the Premier League'' is extracted as the most likely answer to this
question based on the article, even though much of the premise of the question
is incorrect.
Thus, this question fails to penalize the QAFE score of this hallucination.

{\bf QAFE question coverage may be inadequate for some hallucinations.}
Another hallucination which SummaC caught but not QAFE was:
``The team's goalkeeper, Manuel Neuer, was in top form, and the team dominated the game, coming close to scoring several times,''
three questions are generated by QAFE, but none checks whether the team came
close to scoring several times.

On the other hand, at the same threshold, ``NoError'' recall for
QAFE (.710) exceeds SummaC (.580).  Qualitatively, this appears to be
because the sentence-based approach of SummaC prevents it from accepting
multi-hop supports which QAFE can accept by answering multiple questions from
different places.  Overall, SummaC and QAFE have different strengths and
weaknesses, and are best considered together.

\section{Summary Lengths}

\begin{table}[htb]
\footnotesize \resizebox{1\linewidth}{!}{ \parbox{1\linewidth}{
\begin{center}
\begin{tabular}{lcccc}
\hline
& \multicolumn{2}{c}{CNN/DM} & \multicolumn{2}{c}{XSum} \\
Method & Llama & Mistral & Llama & Mistral \\
\hline
Greedy & 175 & 239 & 136 & 124 \\
Nucleus & 267 & 283 & 144 & 147 \\
Beam & 170 & 238 & 151 & 135 \\
\hline
P-CRR & 266 & 217 & 187 & 125 \\
S-CRR & 199 & 245 & 155 & 130 \\
DoLa & 230 & --- & 149 & --- \\
SelfCheckGPT & 171 & 218 & 110 & 152 \\
\hline
Independent & 203 & 224 & 155 & 131 \\
Sample \& Select & 253 & 211 & 113 & 134 \\
\hline
\end{tabular}
\caption{Average lengths of generated summaries in tokens.}
\label{tbl:length}
\end{center}
}}
\end{table}

In Table~\ref{tbl:length}, we compare the lengths of the generated summaries
with each decoding method.  We do not observe consistent trends in
the ordering of the methods with respect to length.

\section{Licenses}

All the article data we require is distributed with FRANK,
which is MIT-licensed.  We also release our crowdworker annotations
and scripts under the MIT license.  Worker ID's are not included,
to preserve anonymity.

\end{document}